\documentclass[letterpaper, 10 pt, conference]{ieeeconf}  

\IEEEoverridecommandlockouts                              

\usepackage{graphicx}      
\usepackage{algorithm} 
\usepackage{algpseudocode} 
\usepackage{varwidth} 
\usepackage{amsfonts}
\usepackage{amsmath}
\usepackage{url}
\usepackage{xcolor}
\newtheorem{theorem}{Theorem}

\usepackage[caption=false,font=footnotesize]{subfig}
\newcommand{\argmin}{\mathop{\mathrm{argmin}}}  
  
\graphicspath{{figures/}}
\title{\LARGE \bf
Efficient Grasp Planning and Execution with Multi-Fingered Hands by Surface Fitting
}

\author{Yongxiang Fan, Masayoshi Tomizuka
\thanks{Yongxiang Fan and Masayoshi Tomizuka are with Department of Mechanical Engineering, 
        University of California, Berkeley, Berkeley, CA 94720, USA
        {\tt\small {yongxiang\_fan, tomizuka}@berkeley.edu}}%
}

\begin{document}
\maketitle
\thispagestyle{empty}
\pagestyle{empty}

\begin{abstract}
This paper introduces a framework to plan grasps with multi-fingered hands. The framework includes a multi-dimensional iterative surface fitting (MDISF) for grasp planning and a grasp trajectory optimization (GTO) for grasp imagination. The MDISF algorithm searches for optimal contact regions and hand configurations by minimizing the collision and surface fitting error, and the GTO algorithm generates optimal finger trajectories to reach the highly ranked grasp configurations and avoid collision with the environment. The proposed grasp planning and imagination framework considers the collision avoidance and the kinematics of the hand-robot system, and is able to plan grasps and trajectories of different categories efficiently with gradient-based methods using the captured point cloud.  The found grasps and trajectories are robust to sensing noises and underlying uncertainties. The effectiveness of the proposed framework is verified by both simulations and experiments. The experimental videos are available at~\cite{website}. 
\end{abstract}

\section{Introduction}
Efficient grasp planning for multi-fingered hands remains a challenging task. First, searching multiple contacts and corresponding hand configurations is a high-dimensional problem. Second, collisions should be considered in both the grasp planning and execution stages. Thirdly, the system should be robust to the uncertainties caused by the imperfect sensing, calibration and actuation. 

Due to the high dimensional state space, majority of the planning researches use sampling based methods~\cite{miller2004graspit,vahrenkamp2018planning,ciocarlie2007dexterous,saut2012efficient,song2018grasp} by employing the precise 3D mesh model of the object. To overcome the curse of dimensionality, the hand poses in~\cite{vahrenkamp2018planning} were sampled around the 
object skeleton. 
To accelerate the inverse kinematics (IK) and collision detection during sampling, the object and fingertip workspace in~\cite{saut2012efficient} were approximated as tree structures. 
Without the gradient information, the sampling-based methods essentially rely on dense sampling and heavy manually-designed heuristics. 
Learning-based methods have been introduced to simplify the heuristics design~\cite{levine2016learning,mahler2017dex}. To reduce the learning dimension, these methods restricted the learned policies to top-down grasps with parallel grippers. 

It is observed that human tends to increase contact surfaces to generate powerful and robust grasps. With larger contact surfaces, the hand provides force/torque from various positions, and the resultant wrench space can resist larger disturbances. 
The idea of matching the hand to the object has been explored by several researchers. 
In~\cite{ciocarlie2007dexterous}, the grasp quality was quantified by the distances between the hand and object. The optimization was rather inefficient without the analytical gradient. 
A data-driven method was proposed in~\cite{li2007data} to find the hand pose that matches features with the object. 
However, it is time consuming to traverse the whole database for feature matching. 
In~\cite{song2018grasp}, the hand-object geometric fitting was achieved by sampling contacts on the object. 
This assignment implicitly maps the kinematics of the hand to the object by manually designed heuristics. The resultant grasps could be a narrow subset of all well-matched grasps.   

To avoid excessive hand-engineering, we proposed a grasp planning method called iterative surface fitting (ISF) in~\cite{fan2018grasp} and increased the efficiency by combining with learning in~\cite{fan2018learning}. 
ISF optimized for both the palm pose and finger displacements by maximizing the surface fitting score. 
However, the previous ISF can only handle the grippers with one degree of freedom (DOF). Moreover, the grasps with collisions were detected and pruned after optimization. The optimize-then-prune operation produced sub-optimal grasps.

Besides planning the desired grasp configurations, the robot should generate proper robot-finger trajectories to execute the grasps. 
The trajectory planning can be extremely expensive in this high dimensional space. 
A method based on rapidly-exploring random tree (RRT) was introduced in~\cite{vahrenkamp2012simultaneous} to plan motion and grasp simultaneously. With the manually designed heuristics,
the RRT dimension was reduced to three. These heuristics defined a potentially narrow and suboptimal subspace for RRT search.
A general trajectory optimization (TrajOpt) algorithm was presented in~\cite{schulman2013finding} using the sequential quadratic programming (SQP)~\cite{boggs1995sequential}. 
Both the RRT and TrajOpt require an object mesh model during the optimization, which is generally absent in the online grasp planning scenario. 

In this paper, we propose a framework to plan and execute grasps with multi-fingered hands. The framework contains both the grasp planning and grasp imagination. The grasp planning searches for optimal grasps by deforming the hand surfaces along its feasible kinematic directions and matching towards the surface of the object, given the assumption that the large matching area produces more stable and powerful grasp. 
With the planned grasp configurations, the grasp imagination ranks the candidate grasps and optimizes the robot-finger trajectories to reach the target grasps given the point cloud representation of the objects under uncertainties. 

The contributions of this paper are as follows. First, the proposed grasp planning is able to find grasps with plausible surface fitting performance efficiently. The average optimization time is 0.40 sec/grasp using the raw point cloud captured by stereo cameras. By optimizing the palm pose and finger joints iteratively, the planning algorithm is able to implement on the hands with multiple DOFs.  
Secondly, the collision is penalized by the gradient-based methods directly, instead of being pruned after optimization as~\cite{fan2018grasp}. 
Furthermore, the proposed method can generate both the power grasps and precision grasps by adjusting the fitting weights of fingertips. 
Finally, the proposed grasp imagination is able to plan collision free finger trajectories in 0.61 sec/grasp with the imperfect point cloud and underlying uncertainties.

The remainder of the paper is as follows. Section~\ref{sec:statement} describes the problem formulation, followed by the proposed grasping framework in Section~\ref{sec:proposed}. The grasp planning and grasp imagination are introduced in Section~\ref{sec:mdisf} and Section~\ref{sec:gi}, respectively. The experimental results on a multi-fingered hand are introduced in Section~\ref{sec:results}. Section~\ref{sec:conclusion} concludes the paper and describes the future work. The experimental videos are available at~\cite{website}.


\section{Problem Statement}
\label{sec:statement}
With the surface contact, the grasp planning for a multi-fingered hand problem can be formulated as:
\begin{subequations}
	\label{eq:general_form}
	\begin{align}
	\max_{R, \boldsymbol{t}, \delta\boldsymbol{q}, \boldsymbol{\mathcal{S}^f}, \boldsymbol{\mathcal{S}^o}} &\  Q(\boldsymbol{\mathcal{S}^f},\boldsymbol{\mathcal{S}^o}) \label{eq1:cost}\\
	s.t. \quad 
	& \boldsymbol{\mathcal{S}^f} \subset \mathcal{T}(\boldsymbol{\partial \mathcal{F}};R,\boldsymbol{t},\delta\boldsymbol{q}), \label{eq1:surface_finger}\\
	& \boldsymbol{\mathcal{S}^o} = NN_{\partial \mathcal{O}} (\boldsymbol{\mathcal{S}^f}), \label{eq1:surface_object}\\
	& dist(\mathcal{T}(\boldsymbol{\partial \mathcal{F}};R,\boldsymbol{t},\delta\boldsymbol{q}), \mathcal{E})\geq 0 \label{eq1:collision}\\
	& \boldsymbol{q}_0 + \delta\boldsymbol{q} \in [\boldsymbol{q}_{\text{min}}, \boldsymbol{q}_{\text{max}}], \label{eq1:constraint2}
	\end{align}
\end{subequations}
where $R\in SO(3), \boldsymbol{t}\in \mathbb{R}^3$ denote the rotation and translation of the hand palm, $\boldsymbol{q}\in \mathbb{R}^{N_{jnt}}$ denotes the joint angle, with ${N_{jnt}}$ representing the number of joints, and $\boldsymbol{q}_0$ and $\delta \boldsymbol{q}$ represent the original and displacement of $\boldsymbol{q}$.  $\boldsymbol{\mathcal{S}^f}=[\mathcal{S}^f_1,...,\mathcal{S}^f_{N_{cnt}}],\boldsymbol{\mathcal{S}^o}=[\mathcal{S}^o_1,...,\mathcal{S}^o_{N_{cnt}}]$ are contact surfaces for all fingers/palms and objects, with $\mathcal{S}^f_i$ and $\mathcal{S}^o_i$ representing the $i$-th contact surface on the finger/palm and object, and $N_{cnt}$ denoting the number of contact surfaces. 
$Q\in \mathbb{R}$ represents the grasp quality related to  $\boldsymbol{\mathcal{S}^f}$ and $\boldsymbol{\mathcal{S}^o}$. Constraints~(\ref{eq1:surface_finger}) shows that $\boldsymbol{\mathcal{S}^f}$ is a subset of the surface transformed from the hand surface $\boldsymbol{\partial\mathcal{F}}$ by $(R,\boldsymbol{t},\delta \boldsymbol{q})$. Constraint~(\ref{eq1:surface_object}) denotes $\boldsymbol{\mathcal{S}^o}$ is computed from the nearest neighbor ($NN$) of $\boldsymbol{\mathcal{S}^f}$ on object surface $\partial \mathcal{O}$.
Constraint~(\ref{eq1:collision}) denotes that the transformed hand surface $\boldsymbol{\partial \mathcal{F}}$ should not collide with the environment $\mathcal{E}$ including the object to be grasped $\partial \mathcal{O}$, ground $\mathcal{G}$ and surrounding objects, and
(\ref{eq1:constraint2}) enforces that $\boldsymbol{q}$ stays in $[\boldsymbol{q}_{\text{min}}, \boldsymbol{q}_{\text{max}}]$. 

Problem~(\ref{eq:general_form}) would be a standard grasp planning problem if all contact surfaces were degenerated into contact points. In general case, however, the problem is challenging to solve by either the sampling based methods or the gradient based methods considering the inverse kinematics (IK) and collision detection with the objects of complex shapes. 

We observe that human tends to match the contact surfaces before grasping in order to expand the grasp wrench space (GWS)~\cite{roa2015grasp} exerted on the object and improve the robustness to uncertainties. GWS-based quality, however, is time-consuming and generally not differentiable. To increase the searching efficiency, a geometric grasp quality is selected as a heuristic. 
More concretely, 
\begin{equation}
\label{eq:Q_form}
Q(\boldsymbol{\mathcal{S}^f},\boldsymbol{\mathcal{S}^o}) = -dist(\boldsymbol{\mathcal{S}^f},\boldsymbol{\mathcal{S}^o}).
\end{equation}

With the geometric quality, the grasp planning searches for candidate grasps by minimizing the surface fitting error between the hand contact surface $\boldsymbol{\mathcal{S}^f}$ and object contact surface $\boldsymbol{\mathcal{S}^o}$, after which a GWS-based quality ranking scheme is adopted to select satisfying grasps. 

Based on this quality formulation, this paper introduces a framework to plan and execute the grasps. 
\begin{figure}[t]
	\begin{center}
		\includegraphics[width=3.3in]{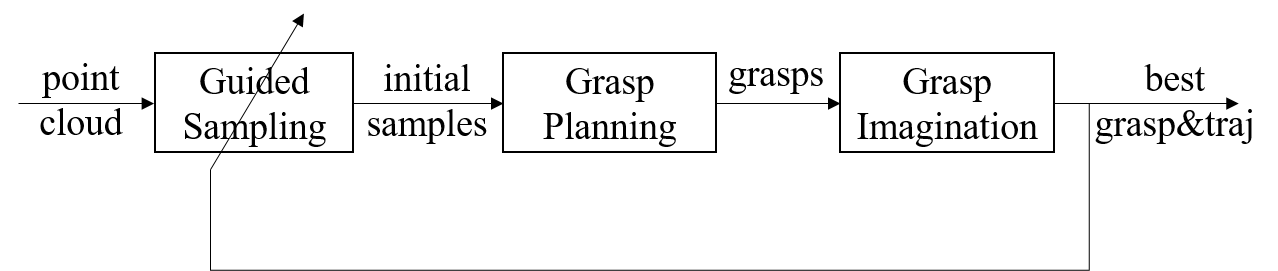}
		\caption{Illustration of the grasp planning and execution framework.}
		\label{fig:framework}
	\end{center}
\end{figure}

\section{The General Planning Framework}
\label{sec:proposed}
This paper introduces an optimization-based framework to plan the desired grasps and optimize robot-finger trajectories to execute the grasps. The framework is illustrated in Fig.~\ref{fig:framework}. It consists of three main blocks: grasp planning by the multi-dimensional iterative surface fitting (MDISF), the grasp imagination by the grasp trajectory optimization (GTO) and the guided sampling. 

Starting with an initial configuration, MDISF fits the hand surface onto the object by optimizing the palm transformation $(R,\boldsymbol{t})$ and joint displacements $\delta\boldsymbol{q}$. 
The registration considers the deformation from the feasible hand motion and the collision between the hand and the environment. 
Compared with non-rigid registration~\cite{myronenko2010point}, MDISF only deforms in the feasible directions of the joint motion.  Compared with the ISF algorithm~\cite{fan2018grasp}, MDISF is able to plan grasps for the hands with multiple DOFs, and the collision between the hand and the object/ground is penalized directly in the optimization. 

The grasp imagination is to evaluate the planned grasps and generate robot-finger trajectories to execute the highly ranked grasps. The GTO algorithm is proposed to avoid collision with the environment and plan optimal finger trajectories using the incomplete point cloud and various types of uncertainties. 

The guided sampling is introduced to avoid being trapped in bad-performed local optima by prioritizing different initial palm placements, so that the regions with smaller fitting errors and better collision avoidance performance can be sampled more often. 
The detail of the guided sampling is in~\cite{fan2018grasp}.  

\section{Multi-Dimensional Iterative Surface Fitting}
\label{sec:mdisf}
\begin{figure}[t]
	\begin{center}
		\includegraphics[width=3in]{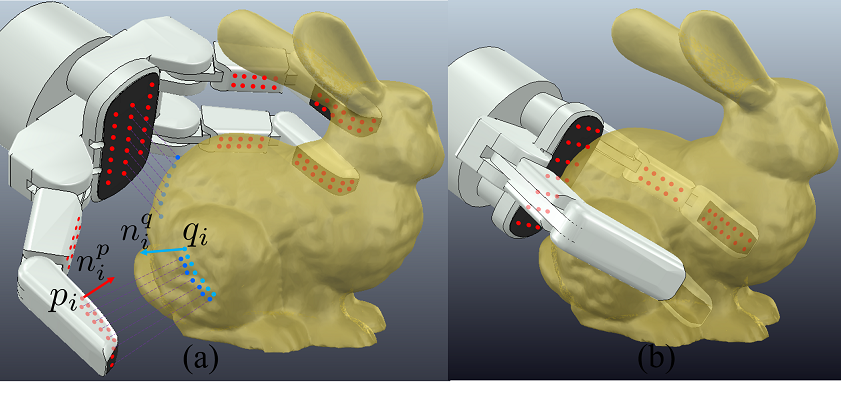}
		\caption{Illustration of the multi-dimensional iterative surface fitting (MDISF) algorithm.}
		\label{fig:ISF_Illustration}
	\end{center}
\end{figure}

With the surface fitting score~(\ref{eq:Q_form}) as the quality, Problem~(\ref{eq:general_form}) can be solved with the proposed multi-dimensional iterative surface fitting (MDISF) algorithm.  MDISF iterates between the correspondence matching and surface fitting, as shown in Fig.~\ref{fig:ISF_Illustration}. To employ the gradient to improve the searching efficiency, the hand surface $\boldsymbol{\partial \mathcal{F}}$ is discretized into points $\{p_i, n_i^p\}_{i=1}^{N_p}$ and bounding boxes $\{\mathcal{B}_k\}_{k=1}^{N_b}$, where $p_i\in \mathbb{R}^3$ and $n_i^p\in \mathbb{S}^2$ represent the point position and normal vector pointing outward, and $N_p$ and $N_b$ are the total number of hand surface points and boxes to cover the surface, as shown in Fig.~\ref{fig:ISF_Illustration}(a). Only the front surface is sampled for simplification. Similarly, the object surface $\partial \mathcal{O}$ is discretized into points $\{q_i, n_i^q\}_{i=1}^{N_q}$, where $q_i\in \mathbb{R}^3$ and $n_i^q\in \mathbb{S}^2$ represent the point position and normal vector pointing outward, and $N_q$ is the total number of points on the object point cloud. 

The correspondence matching finds the paired points $\{q_i\in\mathbb{R}^3, n_i^q\in\mathbb{S}^2\}_{i\in\mathcal{I}}$ on object point cloud by the nearest neighbor search with duplicate/outlier removal~\cite{zinsser2003refined}. 
The surface fitting minimizes the distance between the point pairs $\{p_i, n_i^p\}_{i\in\mathcal{I}}$ and $\{q_i, n_i^q\}_{i\in\mathcal{I}}$, as shown in Fig.~\ref{fig:ISF_Illustration}(b). With the point representation of the surfaces, the surface fitting error $E_{fit}$ is re-formulated as

\begin{equation}
\label{eq:fitting_error}
\begin{aligned}
E_{fit}(R,\boldsymbol{t},\delta \boldsymbol{q})  =
\sum_{i\in\mathcal{I}}&\left((\bar{p}_{i}-q_{i})^Tn_{i}^q\right)^2  + \\&\alpha^2((Re^{(\mathcal{J}_i^w\delta\boldsymbol{q})\string^} n_i^p)\cdot n_i^q +1)^2, 
\end{aligned}
\end{equation}
where $\bar{p}_{i} = Rp_{i} + \boldsymbol{t} +  {R\mathcal{J}_i^v(\boldsymbol{q})\delta \boldsymbol{q}}$
describes the hand surface point after the palm transformation and finger displacement, and $\mathcal{J}_i^v(\boldsymbol{q})$ and $\mathcal{J}_i^w(\boldsymbol{q})$ are respectively translational and rotational Jacobian matrices at the point $p_i$ with the joint $\boldsymbol{q}$. The first term describes the point distance projected to the normal direction of the object surface. This point-to-plane distance is broadly used in ICP~\cite{rusinkiewicz2001efficient} to allow sliding on flat surfaces, so that the algorithm is not sensitive to incomplete point clouds. The second term describes the alignment of the normal vectors. $\alpha$ is to balance the scale of normal alignment.  

With the current correspondence matching and fitting error representation in~(\ref{eq:fitting_error}), the Problem~(\ref{eq:general_form}) becomes:
\begin{subequations}
	\label{eq3:overall}
	\begin{align}
	\min_{R, \boldsymbol{t}, \delta \boldsymbol{q}} &\  E_{fit}(R,\boldsymbol{t},\delta \boldsymbol{q}) \label{eq3:cost}\\
	s.t. \quad 
	& dist\left(\mathcal{T}(\{\mathcal{B}_k\}_{k=1}^{N_b};R,\boldsymbol{t},\delta\boldsymbol{q}), \mathcal{E}\right)\geq 0, \label{eq3:collision}\\
	& \delta \boldsymbol{q} + \boldsymbol{q}_0 \in [\boldsymbol{q}_\text{min}, \boldsymbol{q}_\text{max}], \label{eq3:surface_finger}
	\end{align}
\end{subequations}
where~(\ref{eq3:collision}) represents the collision between the environment $\mathcal{E}$ and the bounding boxes of the hand $\{\mathcal{B}_k\}_{k=1}^{N_b}$.  
Equation~(\ref{eq3:overall}) is a non-convex programming due to the coupling term $R$ and $\delta \boldsymbol{q}$. 

\subsection{Collision Handling} 
\begin{figure}[t]
	\begin{center}
		\includegraphics[width=1.7in]{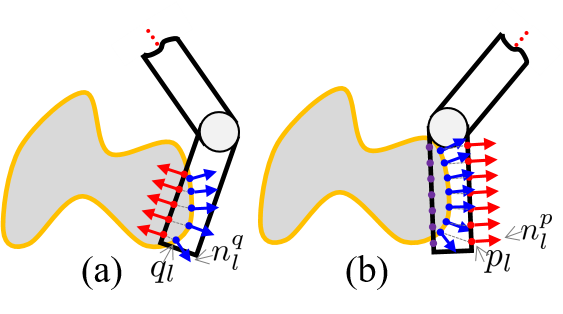}
		\caption{Illustration of different collision types. (a) shows the inner side collision while (b) shows the outer side collision. The sign of $\sum_{l\in\mathcal{L}}n_l^p\cdot n_l^q$ is used for detect the collision type, and $\sum_{l\in\mathcal{L}}n_l^p\cdot n_l^q\leq0$ means inner side contact. For the outer side contact, $p_l$ is replaced by the points on the back (purple dots). }
		\label{fig:collision_type}
	\end{center}
\end{figure}

To address the collision term, we employ the point representation of the  environment and check the inclusion of the points in $\{\mathcal{B}_k\}_{k=1}^{N_b}$. 
Penalty method~\cite{luenberger1984linear} is introduced to avoid collision and ensure that the hand can move smoothly in the space occupied by the object.  More concretely, the collision error is formulated as:
\begin{equation}
\label{eq4:collision_error}
E_{col}(R,\boldsymbol{t},\delta\boldsymbol{q}) =
\sum_{l\in\mathcal{L}_o}\|\bar{p}_{l} - q_{l}\|_2^2, 
\end{equation}
where $\{q_{l}\}_{l\in\mathcal{L}}$ denotes the environment points that are in collision with the bounding boxes.  $\{p_{l}\}_{l\in\mathcal{L}}$ denotes the corresponding points on the box front or back surfaces, and $\bar{p}_{l} = Rp_{l} + \boldsymbol{t} + R\mathcal{J}_{l}^v(\boldsymbol{q})\delta \boldsymbol{q}$. 
To ensure that~(\ref{eq4:collision_error}) reduces all types of collision, we choose the front or the back surfaces that $q_{l}$ paired with, as shown in Fig.~\ref{fig:collision_type}.

With the penalty method, the surface fitting~(\ref{eq3:overall}) becomes
\begin{subequations}
	\label{eq6:overall}
	\begin{align}
	\min_{R, \boldsymbol{t}, \delta \boldsymbol{q}} &\  E(R,\boldsymbol{t},\delta \boldsymbol{q})\label{eq6:cost}\\
	s.t. \quad 
	& \delta \boldsymbol{q} + \boldsymbol{q}_0 \in [\boldsymbol{q}_\text{min}, \boldsymbol{q}_\text{max}], \label{eq6:surface_finger}
	\end{align}
\end{subequations}
where $E(R,\boldsymbol{t},\delta \boldsymbol{q})= E_{fit}(R,\boldsymbol{t},\delta \boldsymbol{q}) + w^2 E_{col}(R,\boldsymbol{t},\delta\boldsymbol{q})$ represents the overall error during the surface fitting of the current correspondence. $w$ denotes the penalty weight of the collision.  

\subsection{Iterative Palm Finger Optimization (IPFO)}
Problem~(\ref{eq6:overall}) is a discretization of~(\ref{eq:general_form}) under the current correspondence and collision penalty, and is solved by the iterative palm finger optimization (IPFO) revised from~\cite{fan2018grasp}. 
The IPFO algorithm iteratively optimizes the palm transformation $(R,\boldsymbol{t})$ and the finger displacements  $\delta\boldsymbol{q}$. 
\subsubsection{Palm Optimization}
The palm optimization searches for optimal $(R, \boldsymbol{t})$ by fixing the finger joint configuration:
\begin{equation}
\label{eq:palm_optimization}
\min_{R, \boldsymbol{t}} E(R,\boldsymbol{t},0) = \min_{x}\|Ax - b\|^2,
\end{equation}
where 
$x = [r^T,\boldsymbol{t}^T]^T\in \mathbb{R}^6$ is a local parameterization of the palm transformation, and $r\in \mathbb{R}^3$ is the axis-angle vector to approximate $R$ in small rotation angle assumption, i.e. $R\approx I + \hat{r}$, where $\hat{\bullet}$ is a skew-symmetric representation of cross product. 
%
Equation~(\ref{eq:palm_optimization}) is a least squares problem and is solved analytically. 
\subsubsection{Finger Optimization}
The finger optimization fixes the palm transformation $(R^*,\boldsymbol{t}^*)$ and searches for optimal finger displacements $\delta \boldsymbol{q}$: 
\begin{subequations}
	\label{eq:finger_optimization}
	\begin{align}
	\min_{\delta \boldsymbol{q}}  &\ E(R^*,\boldsymbol{t}^*,\delta\boldsymbol{q}) = \min_{\delta\boldsymbol{q}}\|C\delta\boldsymbol{q} - d\|^2\\
	s.t. \quad 
	& \delta \boldsymbol{q} + \boldsymbol{q} \in [\boldsymbol{q}_\text{min}, \boldsymbol{q}_\text{max}]. 
	\end{align}
\end{subequations}

Equation~(\ref{eq:finger_optimization}) is a least squares problem with box constraints and can be solved by any constrained least squares solver. 

IPFO feeds as inputs $\boldsymbol{\partial\mathcal{F}}$ represented by $\{p_i, n_i^p\}_{i=1}^{N_p}$, $\partial \mathcal{O}$ represented by $\{q_i, n_i^q\}_{i=1}^{N_q}$,  environment $\mathcal{E}$, the fitting indices $\mathcal{I}$ and collision indices $\mathcal{L}$. The corresponding points for fitting and collision are then sampled for palm optimization and finger optimization. The hand surface and hand configuration are updated after the current iteration. IPFO terminates once the error reduction is less than threshold $\Delta$. IPFO returns the optimal transformation $R,\boldsymbol{t}$, finger displacement $\delta\boldsymbol{q}$ and the updated hand surface $\boldsymbol{\partial \mathcal{F}}$. 

\begin{theorem}
\textit{The IPFO algorithm converges to a local optimum of Problem~(\ref{eq6:overall}). }

\textbf{\proof} The convergence of IPFO is proved based on the global convergence theorem~\cite{luenberger1984linear}. First,  $R,\boldsymbol{t}, \delta\boldsymbol{q} \in D=SE(3)\times \mathbb{R}^{N_{jnt}}$ is a compact set. Second, the function $E(R,\boldsymbol{t},\delta\boldsymbol{q})$ in~(\ref{eq6:overall}) is a continuous function. With the construction of IPFO, we claim that the function $E(R,\boldsymbol{t},\delta\boldsymbol{q})$ is decent since $E(R^*,\boldsymbol{t}^*,\delta\boldsymbol{q}^*)<E(R,\boldsymbol{t},\delta\boldsymbol{q})$ when outside of the solution set. Lastly, the IPFO algorithm composited by the palm optimization $\mathcal{PO}$ and finger optimization $\mathcal{FO}$ is a closed mapping, since $\mathcal{PO}$ is continuous and point-to-point, and $\mathcal{FO}$ is closed in $\mathcal{PO}(R,\boldsymbol{t},\delta\boldsymbol{q})$. 
Therefore, IPFO converges to a local optimum under the current correspondence. 
$\left[\textbf{End of Proof}\right]$
\end{theorem}

\subsection{Fitting Weights Reshaping}
\begin{figure}[t]
	\begin{center}
		\includegraphics[width=2.7in]{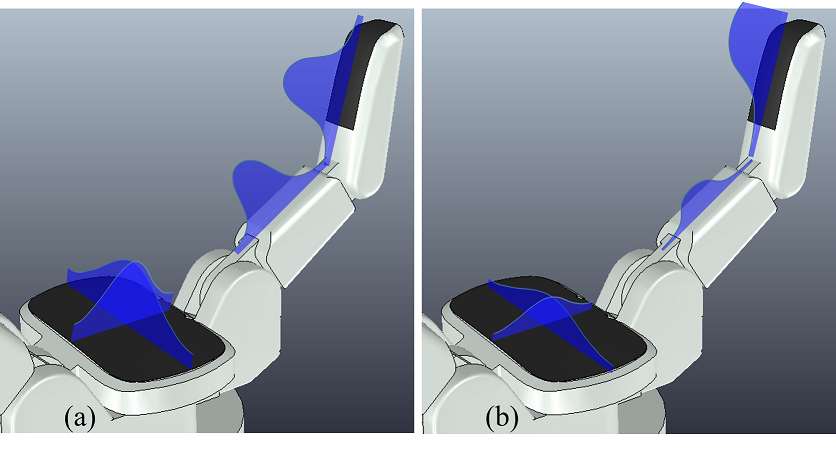}
		\caption{Weights reshaping for (a) power grasp and (b) precision grasp.}
		\label{fig:gaussian_weights}
	\end{center}
\end{figure}

The current MDISF algorithm assumes that all points have equivalent importance. With this assumption, MDISF may produce unsatisfying power grasps which either prevent the hand from closing fingers if it matches to the region close to hinge, or easily collide with the ground if the object is flat. 
To generate natural power grasps, we shape the weights of points on different regions of the hand surface, as shown in Fig.~\ref{fig:gaussian_weights}(a). 
With this shaping, the central regions of palms and links are emphasized since these regions have better robustness to uncertainties and allow large-scale joint motion.  
To produce precision grasps for flat objects, we emphasize the fitting of points on fingertips, as shown in Fig.~\ref{fig:gaussian_weights}(b). 

Similar to~\cite{fan2018grasp}, Gaussian distributions are used to shape the surface points. Despite the variations of pad material and link surface, Gaussian assumption is usually sufficient to balance the offline manual effort and the online searching accuracy based on the grasping results with Barrett hand and Shadow hand. 
The surface fitting with weight shaping is similar to the original one and can be solved by IPFO. The details are omitted for simplicity.

With the IPFO algorithm, MDISF searches optimal hand configuration hierarchically using the multi-resolution pyramid, as shown in Alg.~(\ref{alg:isf}). MDISF iterates between matching the correspondence (Line~\ref{isf:nn}-\ref{isf:collision}) and searching for optimal configuration $R,\boldsymbol{t},\delta \boldsymbol{q}$ with IPFO (Line~\ref{isf:IPFO}). 
\begin{algorithm}[t]
	\caption{Multi-Dimensional Iterative Surface Fitting}\label{alg:isf}
	\begin{algorithmic}[1]
		\State \textbf{Input:} Initial $R_s,\boldsymbol{t}_s, \delta\boldsymbol{q}_s$, $\partial \mathcal{O}, \mathcal{E}$, $\boldsymbol{\partial\mathcal{F}}$, $L, I_0, \epsilon_0$ \label{isf:input}
		\State \textbf{Init:} $\boldsymbol{\partial\mathcal{F}} = \mathcal{T}(\boldsymbol{\partial\mathcal{F}}; R_s, \boldsymbol{t}_s, \delta\boldsymbol{q}_s)$ \label{isf:init}
		\For {$l = L-1, \cdots, 0$} \label{isf:paraymid}
		\State $I_l = I_0/2^l$,  $\epsilon_l= 2^l\epsilon_0$, $e_{prev} \leftarrow \infty$, $\eta \leftarrow 0$, $it \leftarrow 0$
		\While {$\eta \notin [1 - \epsilon_l, 1 + \epsilon_l] $ and $it\texttt{++} < I_l$}
		\State $\mathcal{I} \leftarrow \texttt{filter}(NN_{\partial O}(\texttt{downsample}(\boldsymbol{\partial\mathcal{F}}, 2^l)))$\label{isf:nn}
		\State $\mathcal{L} \leftarrow \texttt{collisioncheck}(\boldsymbol{\partial\mathcal{F}},  \mathcal{E})$ \label{isf:collision}
		\State $\{R, \boldsymbol{t}, \delta \boldsymbol{q}, \boldsymbol{\partial\mathcal{F}}, e\} \leftarrow \textbf{\texttt{IPFO}}(\boldsymbol{\partial\mathcal{F}}, \partial \mathcal{O}, \mathcal{E}, \mathcal{I}, \mathcal{L})$ \label{isf:IPFO}
		\State $\eta \leftarrow e/e_{prev}, e_{prev}\leftarrow e$, $\textbf{Confs}\leftarrow \{R,\boldsymbol{t},\delta\boldsymbol{q}\}$
		\EndWhile
		\EndFor \label{isf:paraymid2} 
		\State \Return $\{ e, \boldsymbol{\partial \mathcal{F}},\textbf{Confs}\}$
	\end{algorithmic}
\end{algorithm}

\section{Grasping Imagination}
\label{sec:gi}
In this section, the found grasps are first ranked based on the proposed quality metric, after which the grasp trajectories are planned to reach the highly ranked grasps. 

\subsection{Grasp Quality Evaluation}
The grasp quality is evaluated based on the grasp wrench space (GWS)~\cite{roa2015grasp}. 
In this paper, GWS $\mathcal{P}$ is constructed by 1) finding contact points by the nearest neighbor of the final hand surface on the object, 2) removing the contacts with large normal alignment errors, 3) extracting the center points and the average normals by K-means, and 4) building $\mathcal{P}$ based on the extracted grasp points and normals using the soft finger model~\cite{murray2017mathematical}. 
With the GWS $\mathcal{P}$, three quality features are calculated. The first feature $Q_{in}$ is a bool type variable and indicates the ability to resist arbitrary small disturbance by checking the inclusion of origin in $\mathcal{P}$. The second feature $Q_{vol}$ indicates the magnitude of disturbance resistance by computing the volume of $\mathcal{P}$. The third feature $Q_{cond}$ indicates the isotropy of the disturbance resistance by the condition number of the $WW^T$, where $W \in \mathbb{R}^{6\times N_p}$ is the vertex matrix of convex hull $\mathcal{P}$. 

The final grasp quality metric $Q_{gsp}$ is represented as:
\begin{equation}
\label{eq:quality_form}
Q_{gsp}=
Q_{vol} + \frac{3}{Q_{cond}} + 11 Q_{in}. 
\end{equation}

The parameters of~(\ref{eq:quality_form}) is obtained by regression using the standard Ferrari-Canny metric~\cite{ferrari1992planning} on 200 grasps from 10 objects. 
Equation~(\ref{eq:quality_form}) is able to rank the found collision-free grasps more efficiently than the Ferrari-Canny metric with comparable accuracy. Compared with Ferrari-Canny metric, the computation time of~(\ref{eq:quality_form}) reduced by 98.77\% from 2.43 secs/grasp to 0.034 sec/grasp. The top-1 score is 70\% and top-3 score is 90\%, out of 20 classes to be ranked. 
Highly ranked grasps are fed for trajectory generation. 

To enhance the robustness of MDISF, the candidate grasps are randomly sampled from a Gaussian distribution to mimic the uncertainties. More concretely, for a candidate grasp $(R_\text{final}, \boldsymbol{t}_\text{final}, \boldsymbol{q}_\text{final})$, we randomly sample $N_t$ times: 
$$(R_t, \boldsymbol{t}_t, \boldsymbol{q}_{t}) \sim \mathcal{N}((R_\text{final},\boldsymbol{t}_\text{final}, \boldsymbol{q}_\text{final}), diag(\Sigma_r, \Sigma_t, \Sigma_q)),$$
where $\Sigma_r\in \mathbb{R}^{3\times 3}, \Sigma_t\in \mathbb{R}^{3\times 3}, \Sigma_{q} \in  \mathbb{R}^{N_{jnt}\times N_{jnt}}$ represent the covariance matrices of the Euler angle of the rotation matrix, translation vector, and joint angles. 
The grasp qualities of different grasp samples are averaged, and the highly ranked grasps are fed for trajectory generation.

\subsection{Grasp Trajectory Optimization}
This paper presents a two-step procedure to plan the robot-finger trajectories. First, the hand keeps half-closed and approaches the pre-grasp pose with~\cite{lin2017real}. The object is represented by its bounding box in this step. The pre-grasp position is defined by lifting the hand $0.3$ m from the final grasp, and the rotation is defined as the closest canonical orientation from the final grasp pose. 
Second, we optimize for the finger trajectories while predefining the palm trajectory by interpolation. The grasp trajectory optimization (GTO) becomes:
\begin{subequations}
	\label{eq:GTO}
	\begin{align}
	\min_{\boldsymbol{q}_1,...,\boldsymbol{q}_S} &\  \sum_{s=1}^{S-1}\|\boldsymbol{q}_{s+1} - \boldsymbol{q}_s\|^2 \label{eq:gto_cost}\\
	s.t. \quad 
	& dist\left(\mathcal{T}(\boldsymbol{\partial \mathcal{F}_s};\boldsymbol{q}_s-\boldsymbol{q}_{s}^0), \mathcal{E}\right)\geq 0,\label{eq:gto_collision}\\
	& |\boldsymbol{q}_{s+1} - \boldsymbol{q}_s| \leq \Delta_q, \quad s=1,...,S-1\\
	& \boldsymbol{q}_{s} \in [\boldsymbol{q}_\text{min}, \boldsymbol{q}_\text{max}], \quad s=1,...,S\\
	& \boldsymbol{q}_{1} = \boldsymbol{q}_\text{pregrasp}, \boldsymbol{q}_{S} = \boldsymbol{q}_\text{final}, \label{eq:gto_prefinal}
	\end{align}
\end{subequations}
where $s$ is the sample index, $S$ is the number of samples on the trajectory, and $\mathcal{T}(\boldsymbol{\partial \mathcal{F}_s};\boldsymbol{q}_s-\boldsymbol{q}_{s}^0)$ denotes the transformed hand surfaces after the joint displacement at the $s$-th sample. Optimization~(\ref{eq:GTO}) is to minimize the total length of the trajectory~(\ref{eq:gto_cost}) from the pre-grasp to final grasp~(\ref{eq:gto_prefinal}) and avoid collision with the environment~(\ref{eq:gto_collision}). 

\begin{figure}[t]
	\begin{center}
		\includegraphics[width=2.2in]{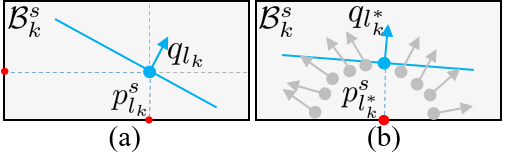}
		\caption{(a) Point-box distance calculation. The closest point $p_{l_k}^s$ is obtained by projection and filtering. (b) The pair with the smallest signed distance ($p_{l_k^*}^s,q_{l_k^*}^s$) is chosen as the critical points for collision avoidance.}
		\label{fig:gto_collision}
	\end{center}
\end{figure}

Similar to MDISF, the collision constraints are penalized in the cost. We adopt the formulation in TrajOpt~\cite{schulman2013finding}:
\begin{equation}
\label{eq:collision_formulation}
col\_term = | d_{safe} - sd(\mathcal{E}, \mathcal{T}(\boldsymbol{\partial \mathcal{F}_s};\boldsymbol{q}_s-\boldsymbol{q}_{s}^0)|^+, 
\end{equation}
where $d_{safe}$ denotes the safety distance, $sd(A,B)$ denotes the signed distance between A and B, and $|x|^+ = \max(x,0)$.

We propose an approach to compute the signed distance $sd(\mathcal{E}, \mathcal{T}(\boldsymbol{\partial \mathcal{F}_s};\boldsymbol{q}_s-\boldsymbol{q}_{s}^0))$ in absence of the 3D mesh and convex decomposition of the environment, as shown in Fig.~\ref{fig:gto_collision}. 
We first inflate the bounding boxes $\{\mathcal{B}_k^s\}_{k=1}^{N_b}$ at sample $s$ by $d_{check}$ and check the inclusion of environment points. For each interior point $q_{l_k}$, we calculate the signed distance by  projecting $q_{l_k}$ to surfaces of $\mathcal{B}_k^s$ and filtering out the points with $(q_{l_k} - p_j)^Tn_l^{q}<0$ for those $q_{l_k}\in \mathcal{B}_k^s$. The closest point to $q_{l_k}$ is denoted as $p_{l_k}^s$, as shown in Fig.~\ref{fig:gto_collision}(a). The point-box signed distance $sd(\mathcal{B}_k^s, q_{l_k})=(p_{l_k}^s - q_{l_k})^Tn_{l_k}^s$ and $n_{l_k}^s$ is a normal vector with direction $q_{l_k} - p_{l_k}^s$ if $q_{l_k} \in  \mathcal{B}_k^s$ or reverse otherwise. 
Therefore, $sd(\mathcal{B}_k^s, \mathcal{E}) = \min_{l_k}sd(\mathcal{B}_k^s, q_{l_k})$ with the critical index $l_k^*=\argmin_{l_k}sd(\mathcal{B}_k^s, q_{l_k})$, as shown in Fig.~\ref{fig:gto_collision}(b). The hand-environment collision indexes $\mathcal{L}_{s}$ is $\{l_k^*\}_{k=1}^{N_b}$.  

With $\mathcal{L}_{s}$, the collision for the $s$-th sample is penalized as: 
\begin{equation}
\begin{aligned}
E_{col,s} = \sum_{l_k^*\in \mathcal{L}_{s}}\left(|d_{safe} - (\bar{p}_{l_k^*}^s - q_{l_k^*})^Tn_{l_k^*}^s|^+\right)^2, 
\end{aligned}
\end{equation}where $\bar{p}_{l_k^*}^s = {p}_{l_k^*}^s + \mathcal{J}_{l_k^*}^v(\boldsymbol{q}_s^0)\delta \boldsymbol{q}_s$. Therefore, GTO~(\ref{eq:GTO}) can be reformulated as: 
\begin{subequations}
	\label{eq:GTO2}
	\begin{align}
	\min_{\delta \boldsymbol{q}_1,...,\delta \boldsymbol{q}_S} &\  \sum_{s=1}^{S-1}\|\boldsymbol{q}_{s+1}^0+\delta\boldsymbol{q}_{s+1}-\boldsymbol{q}_s^0- \delta\boldsymbol{q}_s\|^2 + cE_{col,s} \\
	s.t. \quad 
	& |\boldsymbol{q}_{s+1}^0+\delta \boldsymbol{q}_{s+1} - \boldsymbol{q}_s^0 - \delta \boldsymbol{q}_{s}|_\infty \leq \Delta_q,\\
	& \boldsymbol{q}_{s}^0 + \delta \boldsymbol{q}_{s} \in [\boldsymbol{q}_\text{min}, \boldsymbol{q}_\text{max}], \quad s=1,...,S,\\
	& \delta \boldsymbol{q}_{1} = 0, \delta \boldsymbol{q}_{S} = 0, \\
	& |\delta \boldsymbol{q}_s| < \Delta_{\delta q}. \quad s=1,...,S.
	\end{align}
\end{subequations}
Optimization~(\ref{eq:GTO2}) solves for optimal joint displacements $\{\delta\boldsymbol{q}_s^*\}_{s=1}^S$ using the current joint samples and collided points. The $\{\delta\boldsymbol{q}_s^*\}_{s=1}^S$ then updates joint samples $\boldsymbol{q}_s^0\leftarrow\boldsymbol{q}_s^0+\delta\boldsymbol{q}_s^*$, hand surfaces $\boldsymbol{\partial \mathcal{F}_s}\leftarrow\mathcal{T}(\boldsymbol{\partial \mathcal{F}_s};\delta\boldsymbol{q}_s^*)$, collision penalty $c\leftarrow\mu c$ and indexes $\mathcal{L}_{s}$. Optimization~(\ref{eq:GTO2}) iterates until no collision or reaching the maximum iterations. 


\section{Simulation and Experiment}
\label{sec:results}
This section presents the simulation and experiment. The experimental videos are available at~\cite{website}. 
\subsection{Parameter Lists}
For MDISF, $\alpha=0.03$. $L = 4, I_0 = 200, \epsilon_0 = 0.02$. IPFO used $\Delta = 1e-5, T_{max}= 20$. The power grasp used Gaussian $\Sigma = diag([l/2,w/0.1])$ with mean at the link center, where $l,w$ are the link length and width, and the base weights for palm, proximal and distal link were $0.1, 0.1, 1$. The precision grasp used Gaussian $\Sigma = diag([l/5,w/0.1])$ with mean at the fingertip and base weights $0.01,0.01,1$. 
For robust quality analysis, $N_t = 6+N_{jnt}$, $\Sigma_r = 0.03^2diag([1,1,1])$, $\Sigma_t = 0.003^2diag([1,1,1])$, $\Sigma_q = 0.05^2diag([1,...,1])$. 
As for GTO, $d_{check}=0.03$ m and $d_{safe} = 0.01$ m, $S=30$. For Barrett hand, $\Delta_{\delta q} = [0.2, 0.2, 0.2,0.4]$, $\Delta_{q} = [0.4, 0.4, 0.4,0.4]$, The maximum iterations for (\ref{eq:GTO2}) was 20. The starting collision penalty $c_0$ is $1$ and $\mu$ is $2$. 

\subsection{Simulation Results} 
The simulation was conducted on a desktop with 32GB RAM and 4.0GHz CPU. The grasps were computed by Matlab and visualized by V-REP. The BarrettHand BH8-282  and Shadow Hand were used to test the algorithm. 
\begin{figure}[t]
	\begin{center}
		\includegraphics[width=3.4in]{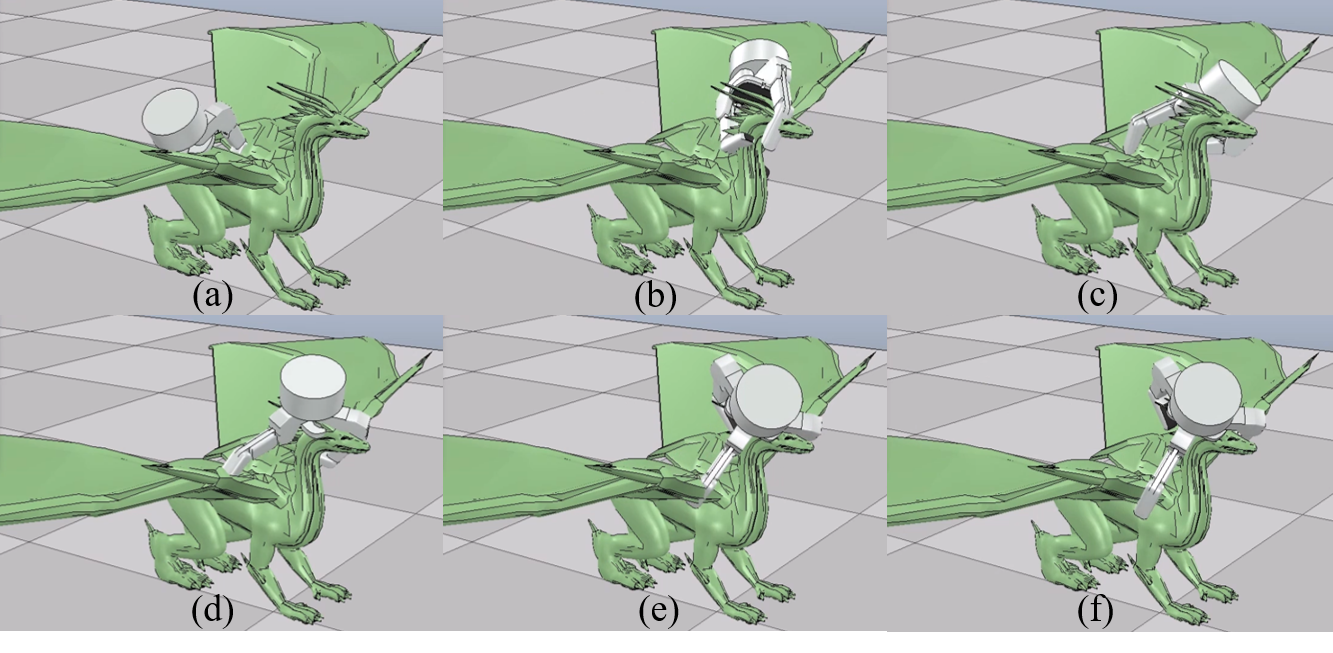}
		\caption{Visualization of MDISF iterations on a dragon object. Snapshot from (a) to (f).}
		\label{fig:ISF_Animation}
	\end{center}
\end{figure}

The visualization of the MDISF iterations are shown in Fig.~\ref{fig:ISF_Animation}. MDISF considers both the collision avoidance and the surface fitting in each IPFO iteration. MDISF started from a random pose around the object (Fig.~\ref{fig:ISF_Animation}(a)) and optimized for palm pose and joint displacements to reduce the fitting error and penalize the collision (Fig.~\ref{fig:ISF_Animation}(b-h)). 

\begin{figure}[t]
	\begin{center}
		\includegraphics[width=3.4in]{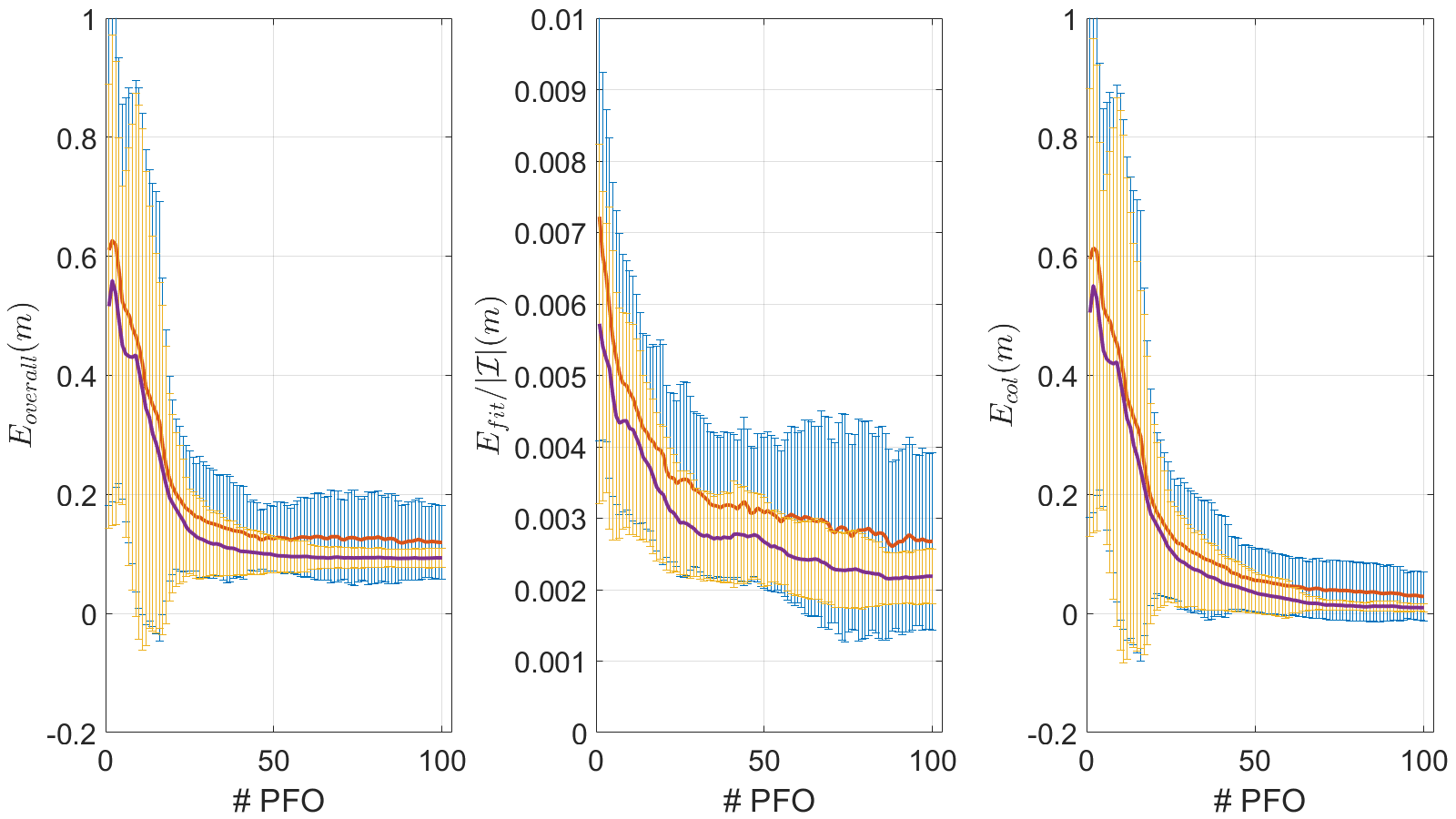}
		\caption{Profile of the error reduction during MDISF. (Left) Overall error. (Middle) Average surface fitting error after outliers/duplicate removal. (Right) Collision error without multiplying the penalty weight. }
		\label{fig:error_reduction}
	\end{center}
\end{figure}

Figure~\ref{fig:error_reduction} shows the error reduction profile to validate that both the average surface fitting error $E_{fit}/|\mathcal{I}|$ and collision cost $E_{col}$ are reduced during MDISF. 
The red and blue plots show the mean errors and the deviations for all samples, while the purple and yellow plots show those for collision-free grasps.  
On average, it took $25.2 \pm 6.3$ IPFOs and $100.2 \pm 34.4$ PFOs to converge. The average fitting error $E_{fit}/|\mathcal{I}|$ reduced from $0.0072 \pm 0.0031$ m to $0.0027 \pm 0.0012$ m, and the absolute $E_{col}$ reduced from $0.5952\pm 0.4342$ m to $0.0287 \pm 0.0410$ m. All statistics were computed based on 50 samples on bunny object shown in Fig.~\ref{fig:ISF_Illustration}. 

\begin{figure}[t]
	\begin{center}
		\includegraphics[width=3.4in]{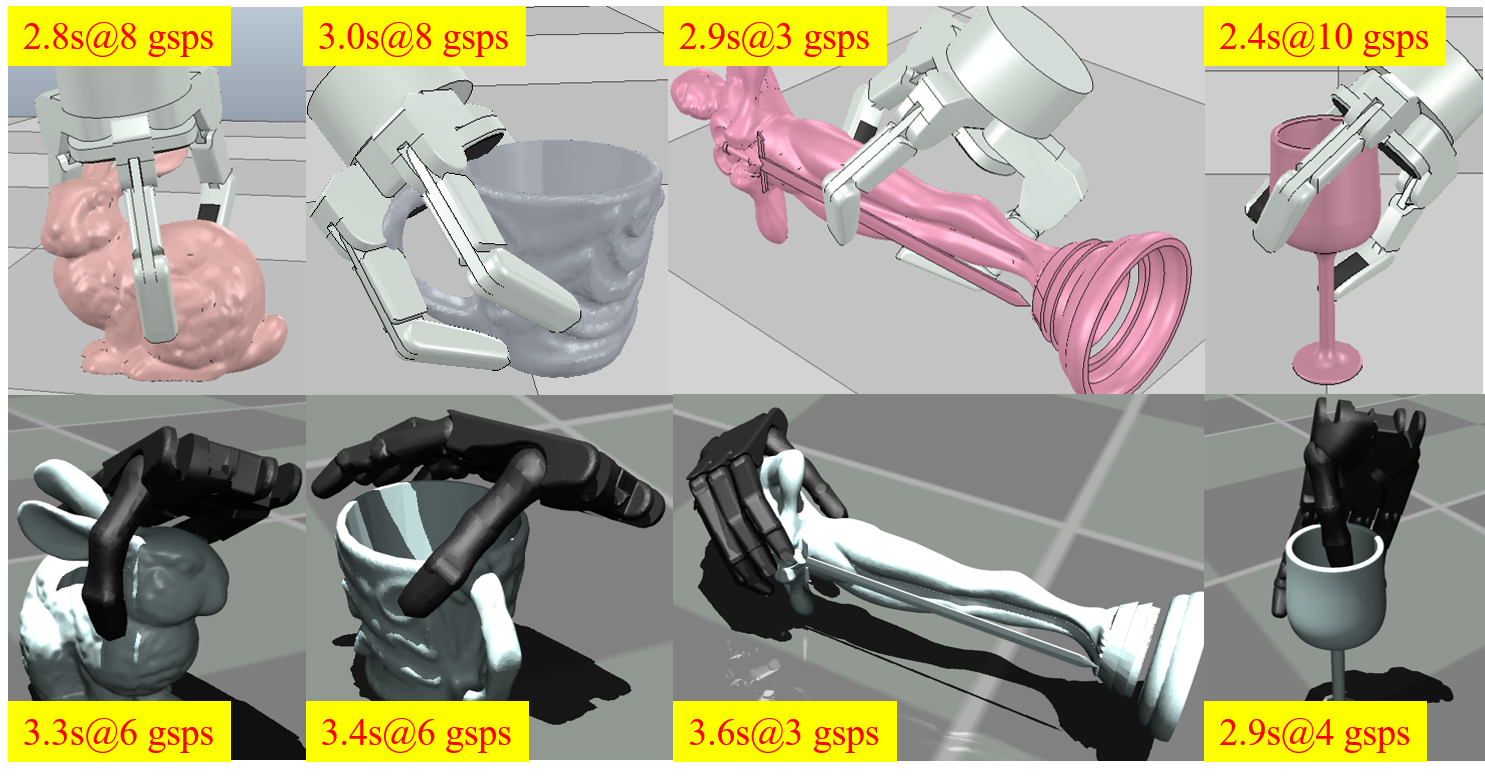}
		\caption{MDISF planning with (Top) Barrett and (Bottom) Shadow hands.}
		\label{fig:sim_vis}
	\end{center}
\end{figure}

Figure~\ref{fig:sim_vis} shows the visualization of the proposed method on four different objects with Barrett Hand and Shadow Hand. The algorithm is able to 1) generate both precision and power grasps, and 2) contact with both the outer and inner surfaces of objects.
Table~\ref{tab:opt_details} shows the quantitative results on ten different objects with the Barrett hand. 
The MDISF algorithm sampled 10 times for each object and returns 6.2 collision free grasps in 2.45 secs. The highest 5 (or the maximum number of collision free grasps found by MDISF) grasps were selected and fed to GTO for trajectory optimization. GTO returned 3.3 collision-free trajectories out of 4.0 grasps in 2.02 secs. The surface fitting error provided a reliable metric for grasp searching, since the majority (6.1/6.2) grasps found by MDISF were force closure (FC). 
As for the grasping with Shadow hand, it took 3.30 secs for MDISF to return 4.75 collision-free grasps in 10 samples.


\begin{table}[t]
		\centering
		\caption{Numerical Results of the Framework with Barrett Hand}
		\label{tab:opt_details}
	\begin{tabular}{l|ll|lll|ll}
		\hline
		\multicolumn{1}{l|}{Object} & \multicolumn{2}{l|}{\begin{tabular}[c]{@{}c@{}} collision-free\#\\ \hline total samples \end{tabular} } & \multicolumn{3}{l|}{Time (secs)} & \multicolumn{2}{l}{Qualities} \\ \hline
				&      ISF& 		GTO&    ISF   &      GTO&     Sum&     FC      &  $Q_{gsp}$         \\\hline\hline
		Bunny &           8/10&     4/5&        2.8&      3.2&    6.0&    7/8&          6.86   \\
		Screwdriver&      9/10&     1/5&       2.2&       3.0&    5.2&    9/9&          9.56 \\
		Gun&           	  2/10&     1/2&       2.5&       1.8&    4.3&    2/2&          -9.41 \\
		Kettle&           8/10&     5/5&       2.1&       2.2&    4.3&    8/8&          3.39 \\ 
		Goblet&              10/10&    5/5&       2.4&       1.7&    4.1&    10/10&        13.66 \\
		Doraemon&         9/10&     5/5&       2.1&       1.6&    3.7&    9/9&          9.27 \\
		Hand&             1/10&     1/1&       2.3&       0.6&    2.9&    1/1&          -12.32 \\
		Banana&           4/10&     3/4&       2.2&       2.5&    4.7&    4/4&          -1.71 \\
		Mug&              8/10&     5/5&       3.0&       2.2&    5.2&    8/8&          8.18\\
		Oscar &           3/10&     3/3&       2.9&       1.4&    4.3&    3/3&          -6.80  \\\hline \hline
		Average&          $\frac{6.2}{10}$&   $\frac{3.3}{4.0}$&     2.45&     2.02  &  4.47 &$\frac{6.1}{6.2}$&      2.068                
	\end{tabular}
\end{table}

\begin{figure}[t]
	\begin{center}
		\includegraphics[width=3in]{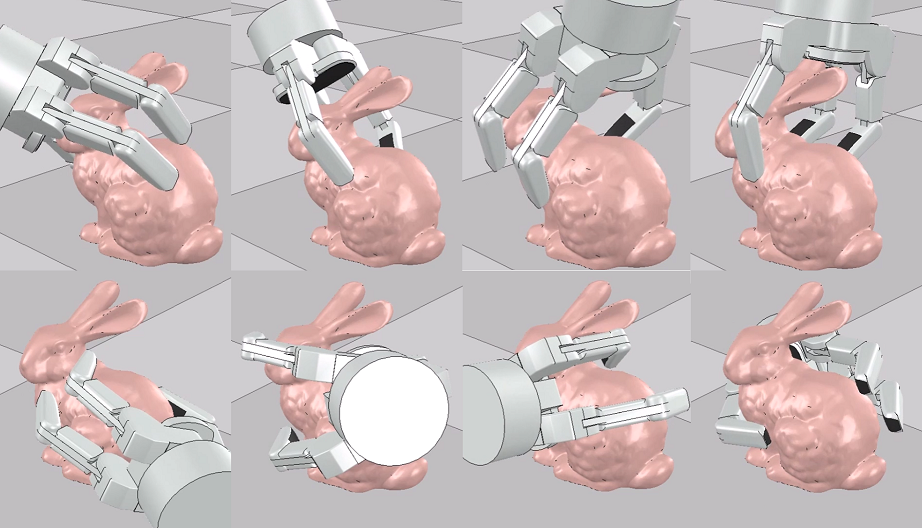}
		\caption{Comparison of (Top) precision grasp mode and (Bottom) power grasp mode of MDISF on Bunny object.}
		\label{fig:precision_and_power}
	\end{center}
\end{figure}

Figure~\ref{fig:precision_and_power} compares the precision grasps and power grasps generated by MDISF. The precision mode and power mode generated 8 and 5 collision-free grasps out of 10 samples on Bunny object, respectively. The hand tended to collide with the object in power grasp mode since the fitting of palm/proximal links were emphasized and hand stayed closer to object, as shown in Fig.~\ref{fig:precision_and_power}(Bottom). 

\begin{figure}[t]
	\begin{center}
		\includegraphics[width=3.3in]{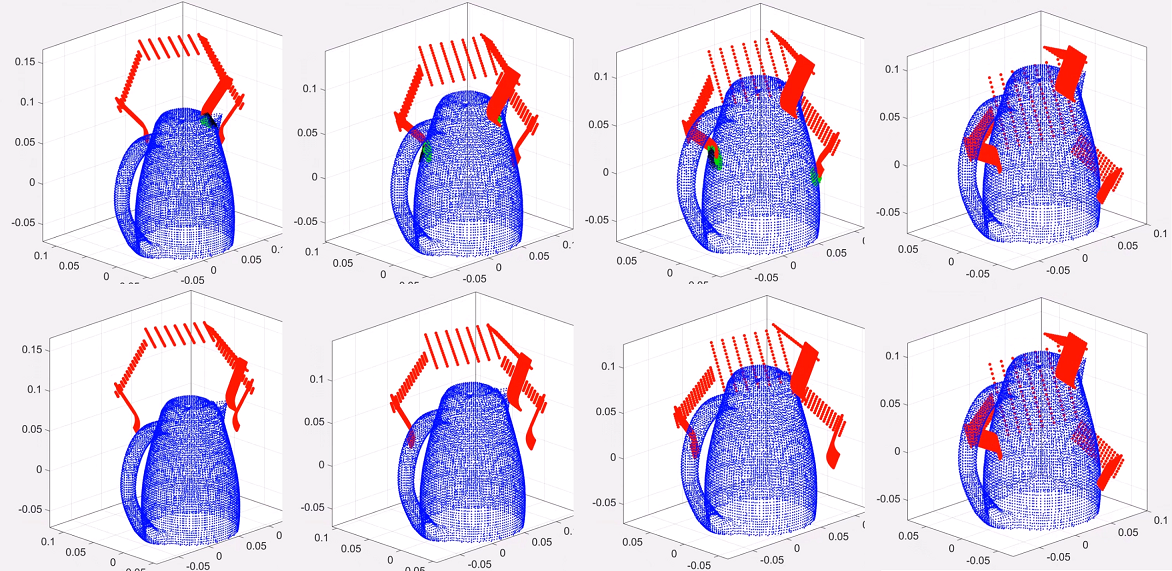}
		\caption{(Top) Trajectory by interpolation. (Bottom) Optimized trajectory by GTO. Green spots are collided regions. Snapshots from left to right.}
		\label{fig:gto_vis}
	\end{center}
\end{figure}

Figure~\ref{fig:gto_vis} shows the result of GTO on kettle object. The trajectory started from the top and reached the desired grasp with one finger in narrow space. The fingers collided with the object when grasping with the predefined finger motion (Fig.~\ref{fig:gto_vis}(Top)). GTO planned finger trajectories to avoid collision and reached the target grasp in narrow space (Fig.~\ref{fig:gto_vis}(Bottom)). 
\begin{figure*}[t]
	\begin{center}
		\includegraphics[width=6.8in]{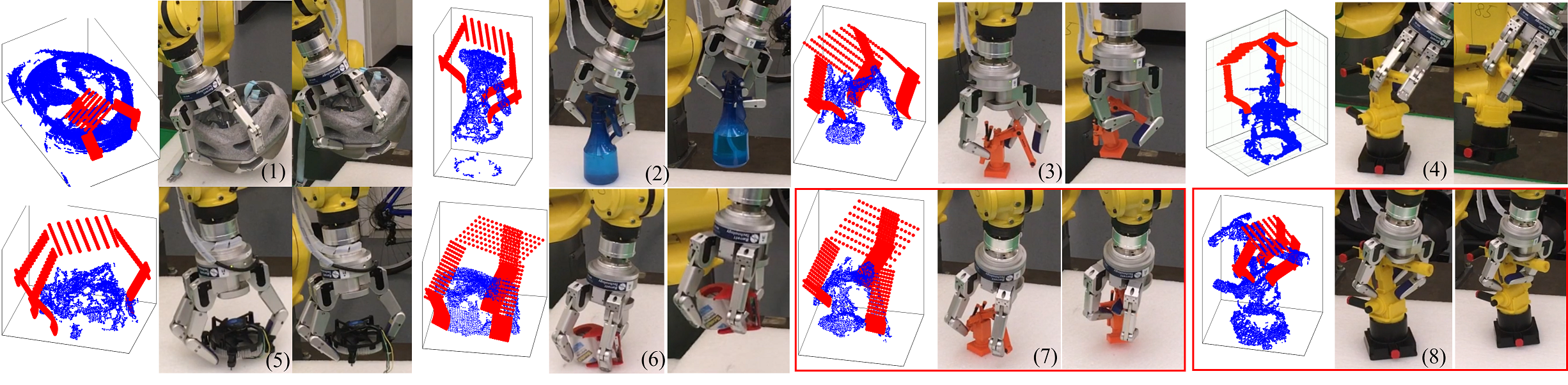}
		\caption{Grasp experiments on 6 objects. (1-6) Successful grasp trials. (7-8) Failed grasp trials. }
		\label{fig:grasp_exp}
	\end{center}
\end{figure*}

\subsection{Comparison with Different Methods}
In this section, the proposed MDISF is compared with two other grasp planning algorithms in~\cite{saut2012efficient} and~\cite{el2013generation}. 

In~\cite{saut2012efficient}, the object and fingertip workspace were approximated as tree structures to accelerate the inverse kinematics (IK) and collision detection during sampling. 
MDISF simultaneously adapts palm and finger joints to higher quality with gradient-based methods, whereas~\cite{saut2012efficient} relies on dense sampling without gradients, thus the grasps may be sampled in low-quality regions. Moreover, MDISF considers collisions with environments (ground and surrounding objects). Lastly, the grasps produced by~\cite{saut2012efficient} are precision grasps, while MDISF can generate both precision grasps and power grasps. The quantitative comparison with a Shadow hand on a mug object is shown in Table~\ref{tab:shadow_comparison}. 
\begin{table}[t]
	\centering
	\caption{Comparison of MDISF and~\cite{saut2012efficient} with Shadow hand on Mug}
	\label{tab:shadow_comparison}
	\begin{tabular}{c|c|c|c}
		items  &   \begin{tabular}[x]{@{}c@{}}Time\\(secs/grasp)\end{tabular}   & \begin{tabular}[x]{@{}c@{}}Additional\\constraints\end{tabular}  & Grasp types\\
		\hline 
		\textbf{MDISF}         & \textbf{0.57}  & \textbf{\begin{tabular}[x]{@{}c@{}}Environmental\\collision\end{tabular}}  &\textbf{ Precision\&power} \\
		\hline
		\cite{saut2012efficient}         & 6.28  & None & Precision only \\
		\hline 
	\end{tabular}
\end{table}

In~\cite{el2013generation}, the object surface is analytically formulated into Gaussian Process (GP). The object modeling takes long time, especially for objects with complex shapes. While a rigorous quality is used in the optimization, the computation is excessively slow for real-time applications. Moreover, the method in~\cite{el2013generation} is unable to model interior surfaces of objects such as bowl, mug and cup. 
The quantitative comparison with the Barrett hand on a toy duck object is shown in Table~\ref{tab:barrett_comparison}. 
For the toy duck object, \cite{el2013generation} selected 241 key points and took 15.1 secs to model the surface. The grasp computation time for one grasp is 12.38 secs. In comparison, the proposed MDISF searches grasps on the raw point cloud and the efficiency is 0.55 sec/grasp. 
\begin{table}[t]
	\centering
	\caption{Comparison of MDISF and~\cite{el2013generation} with Barrett hand on Duck}
	\label{tab:barrett_comparison}
	\begin{tabular}{c|c|c|c}
		items  &   \begin{tabular}[x]{@{}c@{}}Time\\(secs/grasp)\end{tabular}   & \begin{tabular}[x]{@{}c@{}}Additional\\constraints\end{tabular}  & Grasp types\\
		\hline 
		\textbf{MDISF}         & \textbf{0.55}  & \textbf{\begin{tabular}[x]{@{}c@{}}Environmental\\collision\end{tabular}}  &\textbf{ Precision\&power} \\
		\hline
		\cite{el2013generation}         & 15.1+12.38  & None & Precision only  \\
		\hline 
	\end{tabular}
\end{table}

\subsection{Experimental Results}
Figure~\ref{fig:grasp_exp} shows the experimental results using a FANUC LRMate 200iD/7L manipulator and BarrettHand BH8-282 on six objects. The scene was captured by two IDS Ensenso N35 stereo cameras. The observed point cloud and the optimized grasp are shown in the left, and the executed grasp after GTO is shown in the middle. Extra amount of finger motion was executed in order to provide necessary force to clamp the object, as shown in the right. The observed point cloud was incomplete and noisy, and the system also contained uncertainties in calibration ($\sim$3 mm for robot-camera frame alignment), positioning ($\sim$$1^\circ$ TCP-palm alignment, $\sim$$2.0^\circ$ finger joint tracking error) and communication ($\sim$$0.1$ sec robot-hand command synchronization error). The system exhibited certain robustness and was able to plan and execute grasps under the unsatisfying point cloud and various types of uncertainties. We also include two failed grasps to reflect two failure modes. The first failure mode is raised from the perturbation caused by unsynchronized contact (Fig.~\ref{fig:grasp_exp}(7)). The second failure mode is internal disturbance. The proximal link accidentally collided with object during the clamp stage and introduced a large disturbance (Fig.~\ref{fig:grasp_exp}(8)).

\begin{figure*}[t]
	\begin{center}
		\includegraphics[width=6.8in]{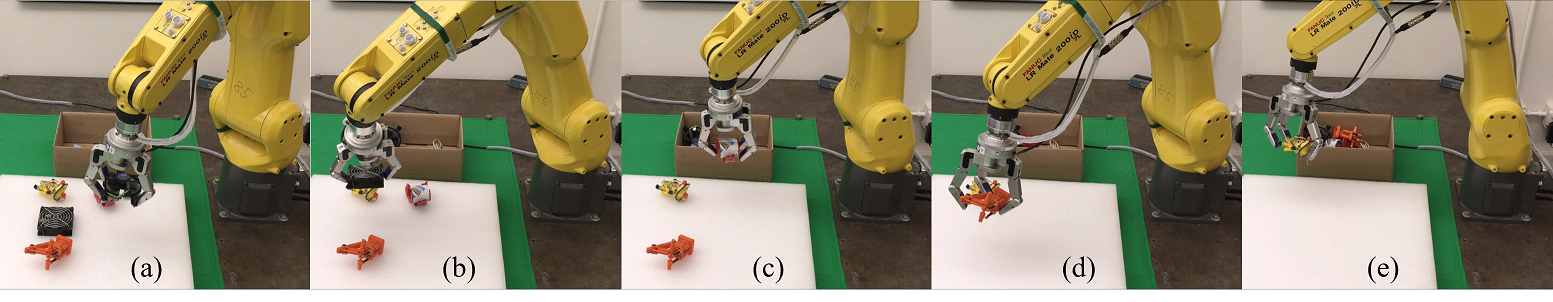}
		\caption{Grasp planning and execution in a clutter environment with success rate 80/97.}
		\label{fig:grasp_exp_clutter}
	\end{center}
\end{figure*}

Finally, Fig.~\ref{fig:grasp_exp_clutter} shows a demonstration of the grasp planning and execution of the proposed framework in clutter environments.  
The Barrett hand may form a large diameter grasp by stretching fingers. To avoid Barrett hand grasping multiple objects in a single grasp, we first segmented the current scene into individual clusters. The largest cluster was then selected for surface fitting by MDISF. The remaining part of the point cloud was regarded as obstacles in the grasp planning and grasp trajectory optimization. A grasp test on 6 objects by 97 grasps shows a 82.47\% success rate.

\section{Conclusion} 
\label{sec:conclusion}
This paper has proposed an efficient framework for grasp generation and execution. The framework includes a multi-dimensional iterative surface fitting (MDISF) and a grasp trajectory optimization (GTO). The MDISF algorithm searches for optimal grasps by minimizing the hand-object fitting error and penalizing the collision, and the GTO algorithm plans finger trajectories for grasp execution with the point cloud representation of the object. The MDISF-GTO exhibits certain robustness to the incomplete/noisy point cloud and various underlying uncertainties. On average, it took 0.40 sec for MDISF to find a collision-free grasp, and took 0.61 sec for GTO to optimize the trajectory to reach the grasp. 

The current implementation clamps objects by simply closing fingers. This may cause uneven force distribution and introduce internal disturbances to the system. Future work will optimize the grasping force~\cite{fan2017real} to have better slippage resistance and disturbance robustness.

\addtolength{\textheight}{-1cm}   
\bibliographystyle{IEEEtran}
\bibliography{references}

\end{document}